\documentclass{paper}
\usepackage[utf8]{inputenc}
\usepackage[T1]{fontenc}    
\usepackage{hyperref}       
\usepackage{url}            
\usepackage{bbm, dsfont}
\usepackage{booktabs}       
\usepackage{amsfonts}       
\usepackage{nicefrac}       
\usepackage{microtype}      
\usepackage{amsmath}        
\usepackage{hyperref}       
\usepackage[title]{appendix}
\usepackage{amsfonts}
\hypersetup{
    filecolor=blue,      
    urlcolor=blue,
}
\usepackage{graphicx}       
\usepackage{pgfplots}       
\usepackage{pgfplotstable}
\usepackage{tikz}           
\usepackage{scalefnt}
\usetikzlibrary{calc}
\usetikzlibrary{decorations.pathreplacing, positioning}
\usetikzlibrary{arrows,patterns, calc,shapes.multipart,chains}
\usetikzlibrary{bending}
\graphicspath{ {./} }
\usepackage[
]{caption}

\pgfdeclarepatternformonly{soft crosshatch}{\pgfqpoint{-1pt}{-1pt}}{\pgfqpoint{4pt}{4pt}}{\pgfqpoint{3pt}{3pt}}%
{
  \pgfsetstrokeopacity{0.3}
  \pgfsetlinewidth{0.4pt}
  \pgfpathmoveto{\pgfqpoint{3.1pt}{0pt}}
  \pgfpathlineto{\pgfqpoint{0pt}{3.1pt}}
  \pgfpathmoveto{\pgfqpoint{0pt}{0pt}}
  \pgfpathlineto{\pgfqpoint{3.1pt}{3.1pt}}
  \pgfusepath{stroke}
}

\definecolor{bluencs}{rgb}{0.0, 0.53, 0.74}
\definecolor{bleudefrance}{rgb}{0.19, 0.55, 0.91}
\definecolor{bluegreen}{rgb}{0.051, 0.596, 0.729}
\definecolor{mediumseagreen}{rgb}{0.24, 0.7, 0.44}
\definecolor{grey}{rgb}{0.5, 0.5, 0.5}
\definecolor{tyrianpurple}{rgb}{0.4, 0.01, 0.24}
\definecolor{darkpastelpurple}{rgb}{0.59, 0.44, 0.84}

\title{Attention Fusion Networks: Combining Behavior and E-mail Content to Improve Customer Support}

\author{
  Stephane Fotso\thanks{Main contributor}, Philip Spanoudes, Benjamin C. Ponedel, Brian Reynoso, Janet Ko\\
  Square, Inc.\\
  San Francisco, CA \\
  \email{\texttt{\{stephane, philips, bponedel,  reynoso, janetko\}@squareup.com}}  \\
}

\begin{document}

\maketitle

\begin{abstract}

Customer support is a central objective at Square as it helps us build and maintain great relationships with our sellers. In order to provide the best experience, we strive to deliver the most accurate and quasi-instantaneous responses to questions regarding our products.

In this work, we introduce the \textbf{Attention Fusion Network} model which combines signals extracted from seller interactions on the Square product ecosystem, along with submitted email questions, to predict the most relevant solution to a seller's inquiry. We show that the innovative combination of two very different data sources that are rarely used together, using state-of-the-art deep learning systems outperforms, candidate models that are trained only on a single source. 


\end{abstract}


\section{Introduction}
Building and maintaining a good relationship with customers is one of the most important outcomes a company can strive for, especially when the customers are entrusting the company with their financing. This is one of the main missions of the Square Capital\footnote{ \href{https://squareup.com/capital}{Square Capital}, the lending arm of Square, Inc., offers short-term business loans to small businesses that use the Square point-of-sale (POS) system for payment processing.} Operations team, which devotes sustained effort to provide the fastest and most tailored response to our sellers, an extension of Square's overall customer-first mindset.

Every year, the team receives tens of thousands of emails from our sellers regarding our products. In order to provide consistent responses to inquiries, Operations Agents (OA) manually classify emails into more than 20 different categories. These include Cost Explanation, Loan Eligibility and Prepayment. They then base their response on the template email associated with the appropriate category. Over the years, inquiry volume has increased in tandem with loan origination volume. The Square Capital Data Science team decided to help alleviate some of the volume of emails by providing solutions to the inquiry before being added to the servicing queue.


To do so, we partnered with our Content and Legal teams to create easy-to-read articles for each individual topic, containing similar information to the email the Operations team would have sent as a reply. Upon submission of their inquiry, the sellers are presented with the relevant articles. If they deem their question to be answered by the content of the articles, they are given the option to not be added to the servicing queue by removing their inquiry.
Moreover, the multifaceted nature of an written inquiry made us realize that a significant number of emails could actually be classified into more than one topic. This gave us the idea to use the model to provide the top three most relevant articles relating to their questions.


Our approach differs from classic text classification as we made the hypothesis that a higher prediction performance could be achieved by combining both
\begin{enumerate}
    \item the content of a seller's email, and
    \item the signals extracted from a seller's interaction on the Square product ecosystem.
\end{enumerate} 
The intuition behind this approach is based on the work of Bogdanova et al. in \cite{bogdanova2017if} and Suggu et al. in \cite{suggu2016deep} in the domain of Community Question Answering. They combine the encoding generated by Recurrent Neural Networks and domain knowledge symbolized by handcrafted features. They observe a significant performance improvement compared to models that only use the text as an input. Here, we hypothesize that a latent interaction exists between a seller's behavior prior to submitting a question, and the content of their email. These interactions are learned through the use of deep-learning models.

Additionally, we consider the fact that the subject of a question is usually dependent on a specific subset of keywords rather than the complete question word set. In other words, we assume that not all words carry the same importance when predicting article relevancy. Thus we introduce an attention mechanism that will compute the relevance of each word so that the model can focus on the most important ones. This approach follows the \href{https://explosion.ai/blog/deep-learning-formula-nlp}{Embed, Encode, Attend, Predict} (EEAP) framework proposed by Matthew Honnibal \cite{eeap}. \newline

In this work, we introduce the \textbf{Attention Fusion Network} model and show that it outperforms candidate models that operate on a single information source (text content or user activity signals). Combining handcrafted features based on user activity and text in the domain of Customer Support was successfully applied by Molino et al. \cite{molino2018cota} at \href{https://eng.uber.com/cota/}{Uber Technologies} via the platform COTA. However, the work performed by our team differs from their approach in the following ways.
\begin{itemize}
    \item The $1^{st}$ version of COTA uses TF-IDF to represent words and we use word embeddings.
    \item The $2^{nd}$ version of COTA uses an Encoder-Decoder architecture (or Encoder-Combiner-Decoder). In contrast, we directly output the predictions without a decoding phase.
    \item We heavily rely on the Attention mechanism and the EEAP framework.
\end{itemize}

 In Section 2, we provide a description of the data. In Section 3, we provide an overview of the type of modeling architectures that we use. In Section 4, we lay out the blueprints of the Attention Fusion Network model. The last section presents performance metrics and compares the results of each individual model as well as the combined solution.
 \newline

\section{Description of the data}
\label{sec:headings}
In this section, we explore the topics that we predict, the type of signals we analyze as well as the text and its vector representation.

\subsection{Topics to Predict}
Out of the 20+ available categories, we only consider the 12 most frequent, as they represent $\sim 80\%$ of all cases that the Operations team handles. If an email's topic doesn't belong to any of these, it is labeled ``Other." 
\begin{itemize}
\item Cost Explanation --- \textit{A general description of the loan repayment process }
\item Decline Follow Up --- \textit{List of the main reasons why a loan application might be declined }
\item Early Payoff --- \textit{Information about the early repayment process}
\item Edit Offer if Already Accepted --- \textit{Questions about modifying an application that the seller has already submitted }
\item Funds ETA --- \textit{An explanation of how and when funds are allocated after the loan application is approved}
\item How to Enroll --- \textit{A simple tutorial on how to enroll for a loan}
\item Increase Options --- \textit{An explanation of whether and how the seller can request more funds than the current loan offer provides }
\item Minimum Repayment Requirement --- \textit{Information about the minimum repayment amount and timetable}
\item Not Eligible for Renewal --- \textit{Questions about why a renewal offer is not yet available}
\item Renewal Eligibility --- \textit{Information about how to become eligible for a renewal offer }
\item No Credit Check --- \textit{An explanation of whether a seller's credit report is used to determine eligibility}
\item Plan Completed --- \textit{Information about the next steps after a seller has repaid their loan in full}
\item Other --- \textit{Any topic that does not belong to the previously listed set of topics}
\end{itemize}


\subsection{Interactions with Square Products}
Thanks to the diverse products that Square offers, we have a comprehensive understanding of how our sellers interact with our ecosystem. 

\subsection{Emails and Word Representations using FastText}
The second type of data at our disposal is the text from the inbound emails themselves. Therefore, we seek to build a model that understands sequences of English words, grammar, and word position in a sentence so as to provide a meaningful numerical representation of the text. \newline

\subsubsection{Pre-processing}
As with most text modeling projects, we perform a pre-processing step on the text before beginning model building.
\begin{itemize}
\item Information that is too specific and doesn't provide any additional meaning to the sentences is detected and removed. This includes dates and dollar amounts --- \textit{Example: ``April 29, 2017" becomes ``this date"}
\item Contractions are replaced by their proper grammatical equivalents --- \textit{Example: ``I’d like a loan" becomes ``I would like a loan"}
\item Finally, for compliance and security reasons, we anonymized PII (Personally identifiable information) data. For instance, phone numbers and email addresses are detected and removed --- \textit{Example: ``john.doe@gmail.com" becomes ``this email address"}.We also encrypted every single word of the text to add an additional layer of security. \newline
\end{itemize}

\subsubsection{Word Embedding}

\textbf{From TF-IDF to Word2Vec} \newline
To represent words in a format that a model can ingest, most text classification models rely on traditional methods such as Bag-Of-Words with TF-IDF \cite{joachims1998text}. But in 2013, the Machine Learning community saw the introduction of \textit{Word2Vec}, a suite of models developed by Mikolov et al.  \cite{mikolov2013efficient}, used for learning vector representations of words or ``word embeddings". This revolutionized NLP research, as Word2Vec provides a robust way to create word embeddings that capture hidden information about a language, like synonyms and semantics, and drastically improve text classification performance.

The main concept behind these models is to generate a representation of a word in a lower-dimensional vector space in which words with similar meanings are \textit{close} to one another.

\textbf{FastText} \newline
Since then, other approaches have been developed to improve the representation mechanisms such as GloVe \cite{pennington2014glove} and Hellinger PCA \cite{lebret2013word}. More recently, Facebook's AI Research (FAIR) group released the library FastText \cite{joulin2016bag}, which represents a significant enhancement over Word2Vec. Instead of directly learning the vector representation of a word like Word2Vec, FastText decomposes a word into a bag of character n-grams, such that the word vector corresponds to the sum of the n-grams vectors. 
It presents the main advantage of being able to generate better embeddings for rare or out-of-vocabulary words, which is particularly useful when the text contains typos or new acronyms. In this work we use the 1 million pre-trained word vectors provided by FastText\footnote{Pre-trained word vectors (1 million) trained on Wikipedia 2017, UMBC webbase corpus and statmt.org news dataset (16B tokens). The vectors can be downloaded at \href{https://fasttext.cc/docs/en/pretrained-vectors.html}{https://fasttext.cc/docs/en/pretrained-vectors.html}} \cite{bojanowski2017enriching} to represent all the words in inbound emails.

\textbf{From word to vector} \newline
Let $S$ be a sequence of $T$ words such that $S = [w_1, w_2, ..., w_T]$, and let $W_{emb} \in \mathbb{R}^{V \times d}$ be the word look-up matrix, where $V$ is the size of the vocabulary and $d$ the dimension of word embedding. Specifically, the $t^{th}$ column of $W_{emb}$ is the vector representation of the $t^{th}$ word in the corpus. Thus, the vector representation of a word is obtained by the dot product
\[ e_t = W_{emb}^{\top} \cdot \hat{w}_t \]
where $\hat{w}_t$ is a one-hot encoding of the word $w_t$ in the corpus. More precisely if $w_t$ represents the $j^{th}$ word in the corpus then $\hat{w}_t[k] = 0$ for all $k \in [1, V]$ such that $k\neq j$, and $\hat{w}_t[j]=1$.

\section{Modeling architectures}

In this section, we provide a description of the models that we use throughout the rest of the paper:
\begin{itemize}
\item Multi-layer Perceptrons (MLP)
\item Recurrent Neural Networks (RNN)
\item The Attention Mechanism
\end{itemize}

\subsection{Multi-layer Perceptrons (MLP)}
Multi-layer Perceptrons or Feedforward Networks \cite{Goodfellow-et-al-2016} are defined by a mapping $f$ between an input $x$, an output $y$ and parameters $\theta$ such that $y = f(x, \theta)$. The parameters $\theta$ are learned so as to provide the best approximation function $f$. These networks are called \textit{multi-layer} because they are represented as a composition of several functions (layers). Specifically,
\[ f(x)= g^{L}(  g^{L-1}( \ldots g^{2}(g^{1}(x)) \ldots ) ) \]
where
\begin{itemize}
\item $x$ is the input
\item $\left\{g^{l}\right\}_{l=1}^{L-1}$ are the hidden layers
\item $g^{L}$ is the output layer. \newline
\end{itemize}

As an input is passed through successive hidden layers a higher level representation of the input vector $x$ is computed. In the specific case of MLPs every hidden layer can be represented by the following formula:
\[ H_l = \sigma \left( W_{l}^{\top} \cdot H_{l-1} + b_l  \right) \]
where
\begin{itemize}
\item $H_l$ is the representation of the $l^{th}$ layer
\item $H_{l-1}$ is the representation of the previous layer
\item $W_l$ and $b_l$ are the parameters to learn
\item $\sigma$ is a nonlinear activation function. \newline
\end{itemize}

Finally, because we are dealing with a multi-class problem we model the probability distribution over $K=13$ different classes. This is accomplished by taking a \textit{Softmax} function, $S(z)$, as the activation function of the output layer $g^L$.  The softmax is defined by $S_i(z) \equiv \frac{e^{z_i}}{\sum_{k}e^{z_k}}$ where $S_i$ for $1\leq i\leq K$ represents the prediction probability of class $i$.


\tikzset{%
   neuron missing/.style={
    draw=none, 
    scale=3.2,
    text height=0.333cm,
    execute at begin node=\color{black}$\hdots$
  },
}

\begin{center}
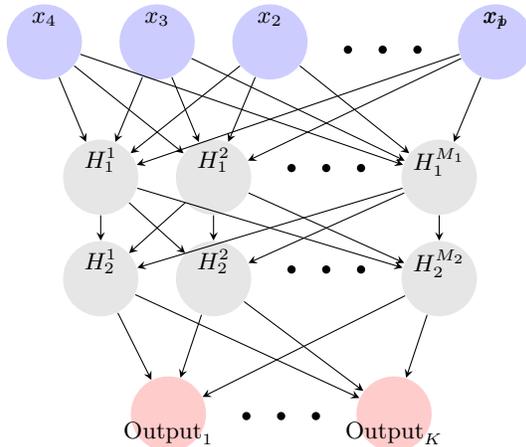
\begin{figure}
\begin{tikzpicture}[x=1.5cm, y=1.5cm, >=stealth]
\foreach \m/\l [count=\x] in {3, 2, 1}{
    \node [circle,fill=blue!20,minimum size=1cm] (input-\m) at (1-\x, 0) {};
}
\foreach \m/\l [count=\x] in {4}{
 \node [circle,fill=blue!20,minimum size=1cm ] (input-\m) at (2, 0) {};
}
\foreach \l [count=\i] in {4,..., 1}
  \node [below] at (input-\i.north) {$x_{\l}$};
\node [below] at (input-4.north) {$x_p$};
\node [neuron missing]  at (1, 0.25) {};
 
\foreach \m/\l [count=\x] in {1, 2}{
  \node [circle,fill=grey!20,minimum size=1cm] (hidden1-\m) at (0.5-\x, -1.2) {};
 }
\foreach \m/\l [count=\x] in {3}
  \node [circle,fill=grey!20,minimum size=1cm ] (hidden1-\m) at (1.5,-1.2) {};
  
 \node [neuron missing]  at (0.5,-0.8) {};
  
\foreach \l [count=\i] in {2, 1}
  \node [below] at (hidden1-\i.north) {$H^{\l}_{1}$};
\node [below] at (hidden1-3.north) {$H^{M_1}_{1}$};

\foreach \m/\l [count=\x] in {1, 2}{
  \node [circle,fill=grey!20,minimum size=1cm] (hidden2-\m) at (0.5-\x, -2.1) {};
 }
\foreach \m/\l [count=\x] in {3}
  \node [circle,fill=grey!20,minimum size=1cm ] (hidden2-\m) at (1.5,-2.1) {};
  
 \node [neuron missing]  at (0.5,-1.7) {};
 
\foreach \l [count=\i] in {2, 1}
  \node [below] at (hidden2-\i.north) {$H^{\l}_{2}$};
\node [below] at (hidden2-3.north) {$H^{M_2}_{2}$};


\foreach \m/\l [count=\x] in {1}
  \node [circle,fill=red!20,minimum size=1cm] (output-\m) at (0.1-\x, -3.3) {};
\foreach \m/\l [count=\x] in {2}
  \node [circle,fill=red!20,minimum size=1cm ] (output-\m) at (1.1,-3.3) {};
  
 \node [neuron missing]  at (0.1,-3) {};
 
\foreach \l [count=\i] in {1}
  \node [below] at (output-\i) {$\text{Output}_{\l}$};
\node [below] at (output-2) {$\text{Output}_{K}$};

\foreach \i in {1,...,4}
  \foreach \j in {1,...,3}
    \draw [->] (input-\i) -- (hidden1-\j);

\foreach \i in {1,...,3}
  \foreach \j in {1,...,3}
    \draw [->] (hidden1-\i) -- (hidden2-\j);

\foreach \i in {1,...,3}
  \foreach \j in {1,...,2}
    \draw [->] (hidden2-\i) -- (output-\j);

\end{tikzpicture}
\captionof{figure}{Representation of a 2-hidden layer Neural Network }
\end{figure}
\centering
\end{center}

\subsection{Recurrent Neural Networks (RNN)}
Recurrent Neural Networks (RNNs)  \cite{rumelhart1986learning} are designed to model sequences of data, like a stream of text. These models recall information about elements seen at previous time steps in the sequence and build connections with future elements.
A RNN can be seen as a discrete dynamical system whose current state is calculated as a function of its previous states, such that:
\[ s_t = F\left( s_{t-1}, x_t, \theta \right) \]
where
\begin{itemize}
\item $s_t$ is the state at step $t$
\item $s_{t-1}$ is the state at step $t-1$
\item $x_t$ is the input at step $t$
\item $\theta$ is a parameter \newline
\end{itemize}

\subsubsection{Long Short Term Memory models (LSTM)}
In practice standard or vanilla RNNs suffer from important shortcomings such as vanishing and exploding gradients. This was established by Bengio et al. in \cite{bengio1994learning} and by Pascanu et al. in \cite{pascanu2013difficulty}. These issues contribute to the rise in popularity of Long Short Term Memory models (LSTM)  \cite{hochreiter1997long} which are designed to avoid them.\newline

Contrary to simple RNNs, the gated LSTM architecture allows the model to regulate the amount of information that is stored or erased. The following equations encode how the gates work ($\sigma$ is the sigmoid function):
\begin{itemize}
\item Input gate: $ i_t = \sigma \left( W_i \cdot \left[ h_{t-1}, x_t \right] + b_i \right) $
\item Forget gate: $ f_t = \sigma \left( W_f \cdot \left[ h_{t-1}, x_t \right] + b_f \right) $
\item Output gate: $ o_t = \sigma \left( W_o \cdot \left[ h_{t-1}, x_t \right] + b_o \right) $
\end{itemize}
where $[.,.]$ is concatenation. 

These gates take into account the previous output of the hidden state and the current input in order to update the \textbf{current memory cell} $c_t$ and the \textbf{current hidden state} $h_t$
  \begin{align*}
     q_t &= \text{tanh}\left( W_q \cdot \left[ h_{t-1}, x_t \right] + b_q \right) \nonumber \\
     c_t &= f_t \odot c_{t-1} + i_t \odot q_t \nonumber\\
     h_t &= o_t \odot \text{tanh}\left( c_t \right)
  \end{align*}
where $\odot$ is element-wise multiplication.

\begin{figure}[h]
\centering
\includegraphics[width=0.4\textwidth]{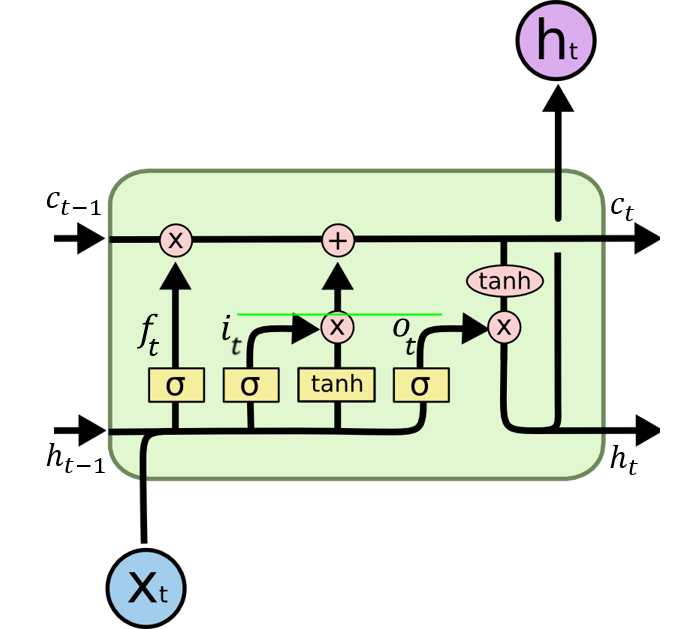}
\caption{Structure of a LSTM cell \cite{UnderstandingLSTM}}
\end{figure}

\subsubsection{Bidirectional LSTM}
In order to increase the chance of capturing long term dependencies in sequential data, Bidirectional LSTMs (BLSTM) \cite{graves2013speech} are often used. A BLSTM  is the concatenation of the hidden states of forward $\overrightarrow{h_t}$ and backward $\overleftarrow{h_t}$ (time-reversed) LSTMs. It provides a representation of the data that depends on both the past and the future such that
\[ h_t = \left[ \overrightarrow{h_t}, \overleftarrow{h_t}\right]. \]
Like the MLP the last layer of the LSTM/BLSTM models used here contain a Softmax activation function.

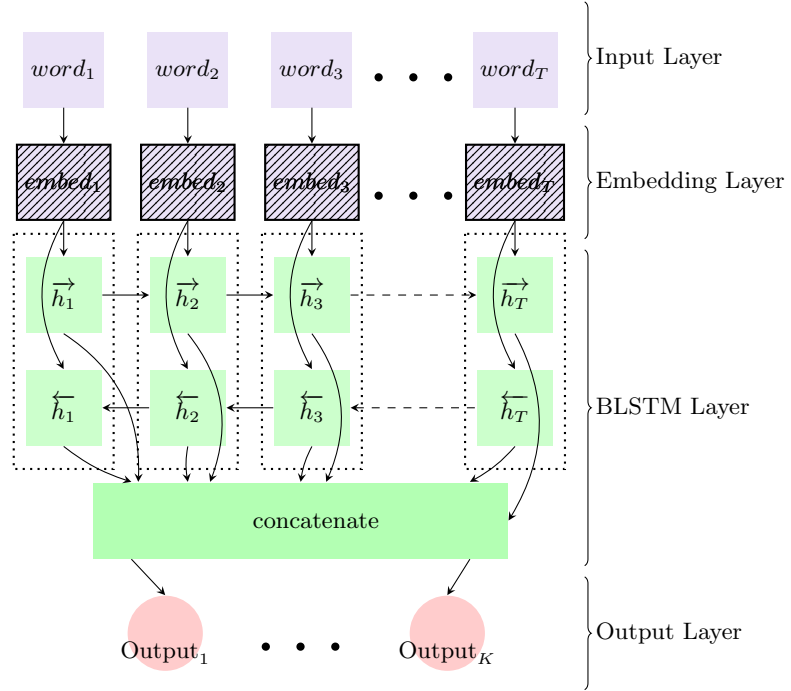
\begin{figure}
\begin{center}
\begin{tikzpicture}[x=1.5cm, y=1.5cm, >=stealth]

\foreach \m/\l [count=\x] in {3, 2, 1}{
    \node [rectangle, fill=darkpastelpurple!20,minimum size=1cm, inner sep=0.1cm] (input-\m) at (0-1.1*\x, 0) {$word_{\l}$};
}
\foreach \m/\l [count=\x] in {T}{
 \node [rectangle,fill=darkpastelpurple!20,minimum size=1cm,inner sep=0.1cm ] (input-\m) at (0.7, 0) {$word_{\l}$};
}
\node [neuron missing]  at (-0.2, 0.25) {};
 
\draw[decoration={brace, raise=5pt},decorate]
  (1.2,0.6) -- node[right=6pt] {Input Layer} (1.2,-0.4) ;

 \foreach \m/\l [count=\x] in {3, 2, 1}{
    \node[minimum size=1cm, fill=darkpastelpurple!20,  inner sep=0.1cm, thick] (embed-\m) at (0-1.1*\x, -1){$embed_{\l}$};
    \node[minimum size=1cm, draw, fill=darkpastelpurple!20,pattern=north east lines, inner sep=0.1cm, thick] (embed-\m) at (0-1.1*\x, -1) {$embed_{\l}$};
}
 \foreach \m/\l [count=\x] in {T}{
    \node[minimum size=1cm, fill=darkpastelpurple!20,  inner sep=0.1cm, thick] (embed-\m) at (0.7, -1){$embed_{\l}$};
    \node[minimum size=1cm, draw, fill=darkpastelpurple!20,pattern=north east lines, inner sep=0.1cm, thick] (embed-\m) at (0.7, -1) {$embed_{\l}$};
}
\node [neuron missing]  at (-0.2, -0.8) {};

\foreach \i in {1,...,3}{
    \draw [->] (input-\i) -- (embed-\i);
}
\draw [->] (input-T) -- (embed-T);

\draw[decoration={brace, raise=5pt},decorate]
  (1.2,-0.5) -- node[right=6pt] {Embedding Layer} (1.2,-1.5) ;

  
 \foreach \m/\l [count=\x] in {3, 2, 1}{
    \node[minimum size=1cm, fill=green!20,  inner sep=0.1cm, thick] (lstm1-\m) at (0-1.1*\x, -2){$\overrightarrow{h_{\l}}$};
}
 \foreach \m/\l [count=\x] in {T}{
    \node[minimum size=1cm, fill=green!20,  inner sep=0.1cm, thick] (lstm1-\m) at (0.7, -2){$\overrightarrow{h_{\l}}$};
}
\draw [->] (lstm1-1) -- (lstm1-2);
\draw [->] (lstm1-2) -- (lstm1-3);
\draw [dashed,->] (lstm1-3) -- (lstm1-T);

 \draw [->] (embed-1.south) --  (lstm1-1.north);
 \draw [->] (embed-2.south) --  (lstm1-2.north);
 \draw [->] (embed-3.south) --  (lstm1-3.north);
 \draw [->] (embed-T.south) --  (lstm1-T.north);

 \foreach \m/\l [count=\x] in {3, 2, 1}{
    \node[minimum size=1cm, fill=green!20,  inner sep=0.1cm, thick] (lstm2-\m) at (-0-1.1*\x, -3){$\overleftarrow{h_{\l}}$};
}
 \foreach \m/\l [count=\x] in {T}{
    \node[minimum size=1cm, fill=green!20,  inner sep=0.1cm, thick] (lstm2-\m) at (0.7, -3){$\overleftarrow{h_{\l}}$};
}
\draw [<-] (lstm2-1) -- (lstm2-2);
\draw [<-] (lstm2-2) -- (lstm2-3);
\draw [dashed,<-] (lstm2-3) -- (lstm2-T);

 \draw[->](embed-1.south) to [bend right]  (lstm2-1.north);
 \draw[->](embed-2.south) to [bend right]  (lstm2-2.north);
 \draw[->](embed-3.south) to [bend right]  (lstm2-3.north);
 \draw[->](embed-T.south) to [bend right]  (lstm2-T.north);

 \foreach \m/\l [count=\x] in { 7, 6, 5, 4, 3, 2, 1}{
    \node[minimum size=1cm, fill=green!30,  inner sep=0.1cm, thick] (concat-\m) at (0.8-0.5*\x, -4){};
}
\node [anchor=west] (note) at (-1.7, -4) { \text{concatenate}};

 \draw[->](lstm1-1.south) to [bend left]     (concat-1);
 \draw[->](lstm1-2.south) to [bend left]  (concat-2);
 \draw[->](lstm1-3.south) to [bend left]  (concat-4);
 \draw[->](lstm1-T.south) to [bend left]   (concat-7.east);

 \draw[->](lstm2-1.south) to [bend right=10]  (concat-1.north);
 \draw[->](lstm2-2.south) to [bend right=10]  (concat-2.north);
 \draw[->](lstm2-3.south) to [bend right=10]  (concat-4.north);
 \draw[->](lstm2-T.south) to [bend left=10]  (concat-7.north);

\draw[decoration={brace, raise=5pt},decorate]
  (1.2,-1.6) -- node[right=6pt] {BLSTM Layer} (1.2,-4.4) ;

 \foreach \m/\l [count=\x] in {3, 2, 1}{
    \draw[thick, dotted] ($(lstm1-\m.north west)+(-0.1,0.2)$)  rectangle ($(lstm2-\m.south east)+(0.1,-0.2)$);
}
\draw[thick, dotted] ($(lstm1-T.north west)+(-0.1,0.2)$)  rectangle ($(lstm2-T.south east)+(0.1,-0.2)$);

\foreach \m/\l [count=\x] in {1}
  \node [circle,fill=red!20,minimum size=1cm] (output-\m) at (-1.4-\x, -5) {};
\foreach \m/\l [count=\x] in {2}
  \node [circle,fill=red!20,minimum size=1cm ] (output-\m) at (0.1,-5) {};

\foreach \l [count=\i] in {1}
  \node [below] at (output-\i) {$\text{Output}_{\l}$};
\node [below] at (output-2) {$\text{Output}_{K}$};
\node [neuron missing]  at (-1.2, -4.8) {};
 \draw[->](concat-1.south) -- (output-1.north);
 \draw[->](concat-7.south) -- (output-2.north);

\draw[decoration={brace, raise=5pt},decorate]
  (1.2,-4.5) -- node[right=6pt] {Output Layer} (1.2,-5.5);

\end{tikzpicture}
\end{center}
\captionof{figure}{Representation of a  Bidirectional LSTM }
\end{figure}

\subsection{Attention Mechanism}
The main intuition behind the attention mechanism is the hypothesis that the essence of a text based message can be traced to a few key words. Thus, computing the relevance of each word so that the model can focus on the most important words might improve performance. The concept was first introduced by Bahdanau et al. in  \cite{bahdanau2014neural} and later adapted by Raffel et al. in \cite{raffel2015feed} as a straightforward reduction mechanism. \newline

Let's consider $h_t$ the current hidden state of a RNN at some time $t$ where $t \in [1,T]$ and $T$ is the sentence length. The weight $\alpha_t$ associated with each $h_t$ is defined to be
\[
 \alpha_t = \frac{\exp(\psi_t)}{\sum_{t=1}^T \exp(\psi_t) } \text{ with }
 \psi_t = \tanh\left(w_{attn}^\top \cdot h_t + b_{attn}\right)
\]
The attention vector $a$ is the weighted average of the sequence of $h_t$, such that $a \equiv \sum_{t=1}^T \alpha_t h_t$. Here, $w_{attn}$ and $b_{attn}$ are the attention parameters to learn.

\begin{figure}
\begin{center}
\begin{tikzpicture}[x=1.5cm, y=1.5cm, >=stealth]

 \foreach \m/\l [count=\x] in {3, 2, 1}{
    \node[minimum size=1cm, fill=green!20,  inner sep=0.1cm, thick] (lstm1-\m) at (0-1.1*\x, 0){$\overrightarrow{h_{\l}}$};
}
 \foreach \m/\l [count=\x] in {T}{
    \node[minimum size=1cm, fill=green!20,  inner sep=0.1cm, thick] (lstm1-\m) at (1, 0){$\overrightarrow{h_{\l}}$};
}
\draw [->] (lstm1-1) -- (lstm1-2);
\draw [->] (lstm1-2) -- (lstm1-3);
\draw [dashed,->] (lstm1-3) -- (lstm1-T);

 \foreach \m/\l [count=\x] in {3, 2, 1}{
    \node[minimum size=1cm, fill=green!20,  inner sep=0.1cm, thick] (lstm2-\m) at (-0-1.1*\x, -1){$\overleftarrow{h_{\l}}$};
}
 \foreach \m/\l [count=\x] in {T}{
    \node[minimum size=1cm, fill=green!20,  inner sep=0.1cm, thick] (lstm2-\m) at (1, -1){$\overleftarrow{h_{\l}}$};
}
\draw [<-] (lstm2-1) -- (lstm2-2);
\draw [<-] (lstm2-2) -- (lstm2-3);
\draw [dashed,<-] (lstm2-3) -- (lstm2-T);

 \foreach \m/\l [count=\x] in {3, 2, 1}{
    \draw[thick, dotted] ($(lstm1-\m.north west)+(-0.1,0.2)$)  rectangle ($(lstm2-\m.south east)+(0.1,-0.2)$) ;
}
\draw[thick, dotted] ($(lstm1-T.north west)+(-0.1,0.2)$)  rectangle ($(lstm2-T.south east)+(0.1,-0.2)$);


\node [circle, fill=grey!10,minimum size=0.5cm] (plus) at (-1, -3) {+};

\foreach \m in {1,...,3, T}{
    \draw [->] ($(lstm2-\m)+(0.1,-0.2)$) -- node[draw,circle,minimum width=0.1cm, label= below:$\alpha_{\m}$] {$\times$} (plus);
}

\node [rectangle,fill=bluegreen!20,minimum size=1cm] (attention) at (-1, -3.5) {attention};
   
\foreach \m/\l [count=\x] in {1}
  \node [circle,fill=red!20,minimum size=1cm] (output-\m) at (-2, -4.5) {};
\foreach \m/\l [count=\x] in {2}
  \node [circle,fill=red!20,minimum size=1cm ] (output-\m) at (0,-4.5) {};

\foreach \l [count=\i] in {1}
  \node [below] at (output-\i) {$\text{Output}_{\l}$};
\node [below] at (output-2) {$\text{Output}_{K}$};
\node [neuron missing]  at (-1, -4.2) {};
 \draw[->](attention.south) -- (output-1.north);
 \draw[->](attention.south) -- (output-2.north);
 
\end{tikzpicture}
\end{center}
\captionof{figure}{Representation of an Attention Layer }

\end{figure}
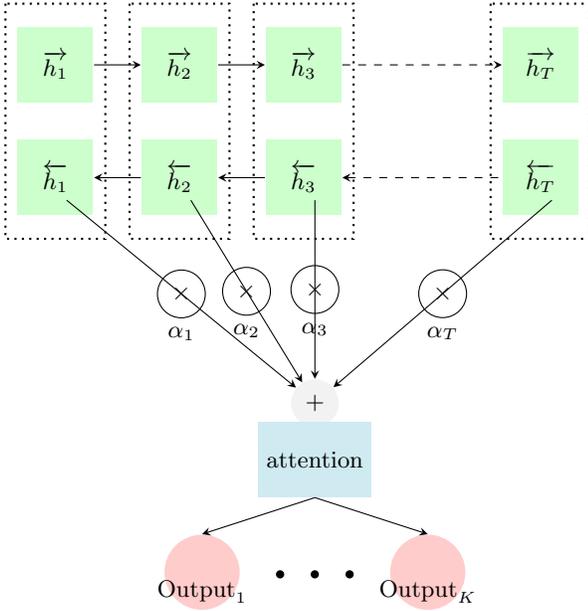

\section{Attention Fusion Networks}
The Attention Fusion Network model consists of concatenating the last hidden layers of the MLP model and of the Attention RNN model.  \newline

\textbf{MLP} \newline
Based on empirical research, building two separate MLP models, one for the continuous handcrafted features and the other for the categorical features encoded as one-hot vectors, provides better results than combining all the features at an early stage. The separated features are then fed to their respective 2-hidden layer neural networks. \newline

\textbf{RNN + Attention} \newline
To build the text model we follow the EEAP framework \cite{sujit2017eeap}. After transforming each word into its vector representation with the FastText vocabulary look-up matrix, we encode the sequences of word embeddings into a sentence matrix by feeding the sequences to a Bidirectional LSTM. We finally use an attention mechanism to summarize the encoded information and make the most important features more salient. \newline

\textbf{Classification} \newline
This concatenated tensor $c$ is fed to a Softmax activation function, $S(z)$, to compute the class probabilities, i.e. $S_i = S(W^{\top} \cdot c_i + b)$.

\begin{center}
\begin{tikzpicture} 

\tikzset{
    >=stealth',
  punktchain2/.style={
    rectangle, 
    rounded corners, 
    draw=black, very thick,
    text width=10em, 
    minimum height=3em, 
    text centered },
  punktchain/.style={
    rectangle, 
    rounded corners, 
    draw=black, very thick,
    text width=5em, 
    minimum height=3em, 
    text centered },
  transparent/.style={
    rectangle, 
    rounded corners, 
    draw=white, very thick,
    text width=5em, 
    minimum height=0.5em, 
    text centered },
}

  [node distance=.8cm] \node[punktchain, fill=blue!20] (mlpn-0) at (-10, 0) {Numerical Features};

  [node distance=.8cm] \node[punktchain, fill=blue!10] (mlpc-0) at (-8, 0) {Categorical Features};

  [node distance=.8cm] \node[punktchain, fill=darkpastelpurple!20] (rnnt-0) at (-6, 0) {Text};

  [node distance=.8cm] \node[punktchain, fill=grey!20] (mlpn-1) at (-10, -3.4) {Fully Connected};
  
  [node distance=.8cm] \node[punktchain, fill=grey!20] (mlpc-1) at (-8, -3.4) {Fully Connected};

  [node distance=.8cm] \node[punktchain,  fill=darkpastelpurple!20] (rnnt-1) at (-6,  -1.7) {Word Embedding};
  
 [node distance=.8cm]  \node[minimum size=1cm, draw, fill=darkpastelpurple!20,pattern=north east lines, inner sep=0.1cm,   text width=5em] (rnnt-2) at (-6,  -1.7) {};

  [node distance=.8cm] \node[punktchain, fill=grey!20] (mlpn-2) at (-10, -1.7*3) {Fully Connected};
  
  [node distance=.8cm] \node[punktchain, fill=grey!20] (mlpc-2) at (-8, -1.7*3) {Fully Connected};

  [node distance=.8cm] \node[punktchain, fill=green!20] (rnnt-3) at (-6, -1.7*2) {BLSTM};

  [node distance=.8cm] \node[punktchain, fill=bluegreen!20] (rnnt-4) at (-6, -1.7*3) {Attention};

\draw[thick, dashed] ($(mlpn-2.north west)+(-0.1,0.2)$)  rectangle ($(rnnt-4.south east)+(0.1,-0.2)$) ;

  [node distance=.8cm] \node[transparent, fill=white!20] (concat) at (-5.9, -1.8*3.4) {Concatenate};

  [node distance=.8cm] \node[punktchain2, fill=red!20] (softmax) at (-8, -1.8*4.2) {Fully Connected with Softmax activation function};

\draw [->] (mlpn-0) -- (mlpn-1);
\draw [->] (mlpn-1) -- (mlpn-2);
\draw [->] ($(mlpc-2.south)+(0,-0.2)$) -- (softmax);

\draw [->] (mlpc-0) -- (mlpc-1);
\draw [->] (mlpc-1) -- (mlpc-2);

\draw [->] (rnnt-0) -- (rnnt-1);
\draw [->] (rnnt-1) -- (rnnt-3);
\draw [->] (rnnt-3) -- (rnnt-4);

\end{tikzpicture}
\end{center}

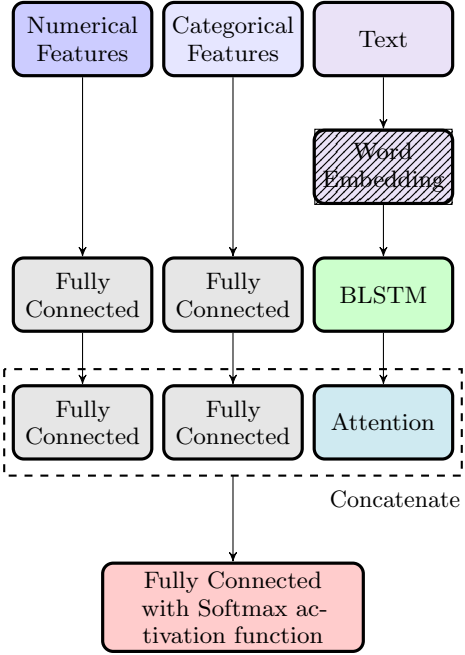
\captionof{figure}{Representation of the Attention Fusion Network model}
\label{fig:hybrid}

\section{Metrics and Results}
In this section, we look at the models' performance. The models were built using Keras \cite{chollet2015keras} with TensorFlow \cite{tensorflow2015-whitepaper}  as backend.

\subsection{Metrics}
Following our Customer Experience team's guidance, it was decided to set the number of displayed articles to 3. Thus we evaluate our model performance with the ``Top 3 Recall" metric and the ``Top 3 Accuracy" metric. The definitions of these metrics can be derived from the general interpretation of the measures given by \cite{zhang2014review}. We introduce the following notation:
\newline

\begin{tabular}{ll}
\hline
Notation & Mathematical Meaning                                                                                   \\ \hline
$n$       & total number of cases to check                                                                                \\
$K$       & total number of classes                                                                               \\
$n_k$     & \vtop{\hbox{\strut number of cases in class $k$ }\hbox{\strut with $k \in  [ 1, K ] $ with $n = \sum_k n_k$ }} \\

$x_i$     & \vtop{\hbox{\strut d-dimensional feature vector for the $i^{th}$ case}\hbox{\strut with with $i \in [ 1, n ] $}}             \\
$Y_i$     & label associated with $x_i$                                                                             \\
$f(x_i)$  & prediction for the $i^{th}$ case \\ \hline
\end{tabular}

\subsubsection{Recall}
Recall is defined as the ratio of cases, for a class $k$, that were correctly predicted by the model. Here, each prediction contains 3 elements such that $f(x_i) = \{ f_{i, 1}, f_{i, 2}, f_{i, 3} \} $, therefore the Top-3 Recall for class $k$ is:
\[ \text{Recall}_3(k) = \frac{1}{n_k} \sum^{n_k} \mathbbm{1}_{Y_i \in \{ f_{i, 1}, f_{i, 2}, f_{i, 3} \}} \cdot \mathbbm{1}_{Y_i = k} \]

\subsubsection{Accuracy}
Accuracy is defined as the proportion of properly predicted results across the total number of cases. Therefore the Top-3 Accuracy is:
\[ \text{Accuracy}_3 = \frac{1}{n} \sum^n \mathbbm{1}_{Y_i \in \{ f_{i, 1}, f_{i, 2}, f_{i, 3} \} } \]

\subsection{Results}

This section goes over the individual performance of each candidate model. Aggregate performance metrics per model can be seen in Table \ref{tab:perf}. Model performances per class are depicted in Figure \ref{fig:classperf}

\subsubsection{MLP model}
The  MLP  model  uses  more than 100 handcrafted features extracted from seller actions within the Square ecosystem. It yields a Top 3 Accuracy of $72\%$. On a per class basis, the model seems to do fairly well on most of the classes. It struggles on classes that require inquiry context. Some of these include ``No Credit Check", ``Edit Offer if Already Accepted" and ``How to Enroll". This is to be expected as it is extremely difficult to predict these sort of classes simply from the way a seller behaves on our systems.

\subsubsection{RNN+Attention model}
The RNN+Attention model uses only the email contents to predict article relevancy. It provides a Top 3 Accuracy of $80\%$. The model performs well across most of the classes, which again is not surprising as it has access to the context of the inquiry. It does underperform on the ``Decline Follow Up" class. We hypothesize that this is due to the model not having access to the system signal that indicates whether a customer's loan application was declined. Along with the fact that these sorts of cases are rare, the RNN+Attention model would require a lot more samples to be able to discriminate these sorts of cases accurately from the context of the inquiry alone.

\subsubsection{Attention Fusion Network}
The Attention Fusion Network model outperforms both the other models on the aggregate, with a Top 3 Accuracy of $87\%$. Furthermore, looking at the model results on a per class basis, we can see that its recall performance is much more balanced across the board. This indicates that the model is able to utilize both information sources and effectively model latent interactions that they share in combination.\newline

\begin{tikzpicture}
\begin{axis}
[
    axis lines*=left,
    hide x axis,
    reverse legend,
    xbar,
    width=4.4cm,
    height=13cm,
    xlabel={},
    symbolic y coords={Early Payoff,Other,Cost Explanation,Minimum Repayment Requirement,How to Enroll,Edit Offer if Already Accepted,Renewal Eligibility,Increase Options,No Credit Check,Funds ETA,Plan Completed,Decline Follow Up,Not  Eligible  for  Renewal},
    ytick=data,
    xmin=0,
    legend style={at={(-1,1)}, anchor=north east}, 
    /pgf/bar width=4.2pt,
    xmax=100, 
    axis line style={draw=none},
    tick style={draw=none},
    ticklabel style={font=\scriptsize}, 
    xticklabel=\pgfmathprintnumber{\tick}\,$\%$,
    nodes near coords={\pgfmathprintnumber\pgfplotspointmeta\%},
    nodes near coords align={horizontal}
    ]

  \addplot [draw=red, fill=red!20] coordinates{ 
        (63,Early Payoff)
        (53,Other)
        (65,Cost Explanation)
        (51,Minimum Repayment Requirement)
        (60,How to Enroll)
        (75,Edit Offer if Already Accepted)
        (75,Renewal Eligibility)
        (77,Increase Options)
        (66,No Credit Check)
        (78,Funds ETA)
        (93,Plan Completed)
        (98,Decline Follow Up)
        (92,Not  Eligible  for  Renewal)
        };

  \addplot [draw=blue, fill=blue!20] coordinates
     {  
        (82,Early Payoff)
        (70,Other)
        (82,Cost Explanation)
        (85,Minimum Repayment Requirement)
        (77,How to Enroll)
        (83,Edit Offer if Already Accepted)
        (85,Renewal Eligibility)
        (85,Increase Options)
        (87,No Credit Check)
        (86,Funds ETA)
        (86,Plan Completed)
        (69,Decline Follow Up)
        (88,Not  Eligible  for  Renewal)
        };

  \addplot
    [draw=black, fill=black!20]
coordinates
     {  
        (79,Early Payoff)
        (79,Other)
        (80,Cost Explanation)
        (80,Minimum Repayment Requirement)
        (83,How to Enroll)
        (87,Edit Offer if Already Accepted)
        (88,Renewal Eligibility)
        (88,Increase Options)
        (91,No Credit Check)
        (92,Funds ETA)
        (92,Plan Completed)
        (94,Decline Follow Up)
        (95,Not  Eligible  for  Renewal)
        };
      \legend{MLP, RNN, Hybrid}

\end{axis}
\end{tikzpicture}
    \captionof{figure}{Comparison of the models performances--- Top-3 Recall by class}
\label{fig:classperf} 

\hfill \break
There are certain cases where the hybrid model under-performs compared to the other two candidate models (i.e ``Minimum Repayment Requirement", ``Cost Explanation" and ``Early Payoff"). We attribute this effect to the performance metric we employ, Top 3 Recall. In fact, the ``Other" class has the largest relative performance increase across all other classes when comparing the Attention Fusion Network to the other two models. This suggests that the effect seen on the aforementioned classes is due to the other two candidate models under-fitting the ``Other" class. 

\begin{table}[!htb]
\begin{center}
\begin{tabular}{ll}
\hline Model                        & Top-3 Accuracy  \\ \hline
     MLP                            & $72\%$          \\
     RNN + Attention                & $80\%$          \\   
     Attention Fusion Network      & $87\%$          \\ \hline
\end{tabular}
\captionof{table}{Comparison of the models performances --- Top-3 Accuracy}
\label{tab:perf}
\end{center}
\end{table}

\section{Conclusion}
In this project, we describe a robust prediction algorithm for the purpose of servicing seller inquiries. To accomplish this, we combine two distinct deep learning models into a hybrid model called the \textbf{Attention Fusion Network} model.
\begin{itemize}
\item The first model only uses the data related to seller interactions with Square products.
\item The second focuses on the text contained in inbound email inquiries. \newline
\end{itemize}

Overall, our solution outperforms both candidate models and shows that complex interactions between customer interaction signals and inquiry context can be leveraged to inform better predictions. \newline

One of the most interesting aspects of this work is uncovering the interaction between textual features and non-text features. In particular, upon combining the two sources, the overall predictive ability is improved substantially. The non-text signals help amplify critical aspects of the text that are impossible to learn without additional context.
\newline

This result suggests that in industrial applications of NLP where a vast amount of text data is not available (in terms of samples) or in cases where the underlying text data cannot predict the outcomes alone; available external signals can be leveraged to increase prediction performance on the task at hand. The Attention Fusion Network introduced here is one attempt to accomplish this goal, as well as to provide the fastest and most tailored response to our sellers. 

\section*{Acknowledgment}
We would like to thank Wafa Soofi, Jacqueline Brosamer, Jochen Bekmann and Tony He, as well as Sara Vera and Jessie Fetterling.

\bibliographystyle{abbrv}
\bibliography{references}

\end{document}